\documentclass{article}

\usepackage[main, final]{neurips_2026}

\usepackage[utf8]{inputenc} 
\usepackage[T1]{fontenc}    
\usepackage{hyperref}       
\usepackage{url}            
\usepackage{booktabs}       
\usepackage{amsfonts}       
\usepackage{nicefrac}       
\usepackage{microtype}      
\usepackage{xcolor}         

\usepackage{array}
\usepackage{xcolor}
\usepackage{colortbl}
\usepackage{makecell}
\usepackage{subcaption}
\usepackage{tabularx}
\usepackage{tcolorbox}
\usepackage{listings}
\usepackage{wrapfig}
\usepackage{algorithm}
\usepackage{algorithmic}
\usepackage{float}
\usepackage{amsmath}
\usepackage{amssymb}
\usepackage{enumitem}
\tcbuselibrary{breakable}

\title{Combinatorial Synthesis: Scaling Code RLVR via Atomic Decomposition and Recombination}

\author{%
  Jiasheng Zheng$^{1,2}$
  \quad
  Boxi Cao$^{1}$
  \quad
  Boxi Yu$^{3}$
  \quad
  Yuzhong Zhang$^{4}$
  \quad
  Jialun Cao$^{5}$\\
  \textbf{Yaojie Lu}$^{1}$
  \quad
  \textbf{Hongyu Lin}$^{1}$
  \quad
  \textbf{Xianpei Han}$^{1}$
  \quad
  \textbf{Le Sun}$^{1}$\\
  $^{1}$Chinese Information Processing Laboratory, Institute of Software, Chinese Academy of Sciences\\
  $^{2}$University of Chinese Academy of Sciences\\
  $^{3}$Lero the Research Ireland Centre for Software, University of Limerick\\
  $^{4}$The Chinese University of Hong Kong, Shenzhen\\
  $^{5}$The Hong Kong University of Science and Technology\\
  \texttt{\{zhengjiasheng2022,caoboxi,luyaojie,hongyu,xianpei,sunle\}@iscas.ac.cn} \\
}

\begin{document}

\maketitle

\begin{abstract}
Reinforcement Learning with Verifiable Rewards (RLVR) has recently emerged as the cornerstone for shaping the remarkable coding abilities of Large Language Models (LLMs).
However, the scalability of RLVR is severely constrained by the scarcity of sufficiently challenging verifiable code tasks that target near the model's edge of competence.
Prior studies often rely on heuristic seed expansions for data synthesis, which severely limits both novelty and difficulty. 
Consequently, the training value of such data fails to scale proportionally with the size of its synthesis.
To this end, we propose \textit{\textbf{A}tomic \textbf{D}ecomposition and \textbf{R}ecombination} (ADR), a novel framework that generates verifiable code tasks via decomposition into atomic elements and controlled recombination, thereby enabling the generation of genuinely novel and challenging verifiable code tasks. 
Experiments and analysis demonstrate that ADR achieves superior originality, difficulty, diversity, and test quality over existing baselines, and consistently delivers greater improvements in code ability across RLVR in diverse downstream domains, including algorithmic programming, tool usage, and data science.
Our work sheds light on a new paradigm for novel code task synthesis and scalable RLVR training~\footnote{Our source code and datasets are available at \url{https://github.com/icip-cas/ADR}}.
\end{abstract}

\section{Introduction}

Reinforcement Learning with Verifiable Rewards (RLVR) has emerged as the cornerstone for shaping the strong coding capabilities of large language models (LLMs)~\cite{jaech2024openai,xu2025towards,zhang2025survey}. 
Leveraging the executability of code, RLVR based on deterministic unit tests can substantially enhance the logical reasoning and code generation abilities of LLMs, enabling their broad adoption in scenarios such as code agents and software development pipelines~\cite{guo2025deepseek,lambert2024tulu,wang2025agents}. 
However, the effectiveness of RLVR critically depends on the availability of large-scale, challenging code tasks equipped with rigorous test cases \cite{wen2025reinforcement}. 
Unfortunately, manually constructing such datasets requires substantial human effort~\cite{villalobos2024position,zhao2025absolute}, making it difficult to scale.
Therefore, data scarcity has become a primary bottleneck limiting the further development of RLVR.

\begin{figure*}[!t]
  \begin{center}
    \centerline{\includegraphics[width=\columnwidth]{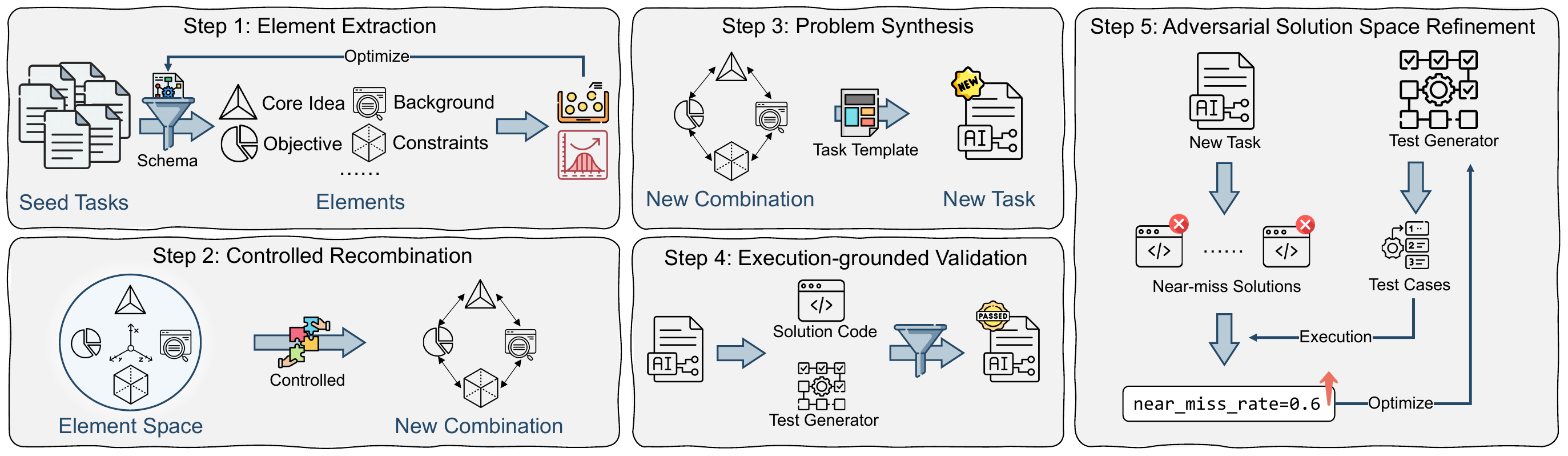}}
    \caption{Overview of the Atomic Decomposition and Recombination (ADR) framework. ADR extracts atomic elements from domain seeds to build an element space, then synthesizes and validates new tasks via controlled recombination and adversarial refinement.}
    \label{fig:main}
  \end{center}
\end{figure*}

An intuitive approach for alleviating this data scarcity is data synthesis. 
Nevertheless, existing methods for synthesizing verifiable code data lag behind the advances observed in pretraining or instruction-tuning data.
A key reason lies in prior findings that \textit{RLVR produces true capability gains only when the tasks are sufficiently challenging and target near the model's edge of competence} \cite{zelikman2022star,liu2025prorl,sun2025rl,zhang2025interplay,yu2025rlpr}, a threshold existing methods fail to reach \cite{huang2025opencoder,wei2024magicoder}. 
In particular, existing approaches predominantly rely on heuristic expansions of existing tasks, such as in-context or recursive prompting \cite{codealpaca,zeng-etal-2025-acecoder,xu-etal-2025-kodcode,luo2024wizardcoder}. 
While these methods increase linguistic diversity, they fail to expand logical diversity or task difficulty, thereby offering limited challenge to the agent’s exploration policy. 
In practice, such superficial expansion restricts agent exploration and leads to premature reward saturation during RLVR.
Therefore, we believe that this bottleneck is not a matter of implementation, but an intrinsic limitation of the heuristic expansion paradigm: by preserving the original compositional structures of seeds, it precludes the generation of genuinely novel logical topologies.

Building on these insights, we propose a novel verifiable code data synthesis framework, \textit{\textbf{A}tomic \textbf{D}ecomposition and \textbf{R}ecombination} (ADR). 
Unlike prior approaches that rely on heuristic seed expansions, ADR constructs data by intersecting orthogonal logical primitives, thereby enabling the generation of genuinely novel and challenging verifiable code tasks. 
Specifically, as illustrated in Figure~\ref{fig:main}, given a small set of domain-specific seed data, we first extract atomic elements via a schema-based, information-theoretically optimized process to form an element space. 
We then synthesize new tasks through controlled recombination, generate solutions and test generators for execution-based validation, and further enhance test quality via adversarial solution-space optimization.
By navigating this combinatorial space, ADR transcends the distributional boundaries of seed data to produce structural intersections unattainable via mere extrapolation.
Notably, ADR’s fully automated design enables rapid adaptation to diverse code tasks such as algorithmic programming, tool usage, and data science with minimal seed data, achieving scalable coverage across both task domains and data scale.

To comprehensively evaluate the quality of the synthesized verifiable data, we introduce a multi-dimensional evaluation taxonomy, including originality, difficulty, diversity, and test quality. 
Results demonstrate that ADR-synthesized data significantly outperforms prior synthesis data (e.g., KodCode~\cite{xu-etal-2025-kodcode} and Educational Instruct~\cite{huang2025opencoder}), across all evaluative dimensions. 
Furthermore, we conduct RLVR training across diverse code domains and base models.
Extensive experimental results reveal that: 
(1) Previous synthetic data methods, constrained by heuristic expansions of real-world data, fail to surpass original data performance; 
(2) ADR-synthesized data consistently yield improvements across various base models and code domains. On LCB-v5, ADR achieves 25.37\% (+9.20\%) on Qwen2.5-Coder-7B, outperforming the best baseline's 22.75\%;
(3) Crucially, unlike prior methods that merely enhance sampling density without improving core reasoning (resulting in only +0.60\% Pass@8 gains), ADR achieves a remarkable +4.79\% Pass@8 improvement, effectively expanding the model's capability frontier.

Our main contributions are summarized as follows:
\begin{enumerate}
    \item We propose the ADR framework, a novel paradigm that shifts from heuristic seed expansion to atomic decomposition and compositional recombination.
    \item We establish a multi-dimensional evaluation taxonomy for verifiable synthetic data that integrates data quality metrics with downstream RLVR performance.
    \item Through extensive experiments on large-scale RLVR, we show the strong generalization ability and practical effectiveness of ADR across multiple code domains.
\end{enumerate}

\section{Related Work}

\subsection{Reinforcement Learning with Verifiable Rewards}

RLVR has emerged as an effective paradigm for eliciting complex reasoning behaviors in LLMs via automatically verifiable signals~\cite{guo2025deepseek,hu2025open,wen2025reinforcement}.
Most prior work focuses on domains with well-defined correctness criteria, particularly math and code~\cite{zeng-etal-2025-acecoder}.
In the math domain, systems such as Kimi K1.5~\cite{team2025kimi}, Tulu 3~\cite{lambert2024tulu}, and SimpleRL-Zoo~\cite{zeng2025simplerl} scale RLVR on datasets like GSM8K and MATH.
In the code domain, RLEF~\cite{gehring2025rlef} use code dataset CodeContests in RLVR.
Beyond strictly verifiable tasks, recent work extends RLVR to less structured domains using soft reward signals, including generative scoring in unstructured answers scenarios~\cite{su2025crossing} and rubric-based rewards in medical and scientific scenarios~\cite{gunjal2025rubrics}.
Despite these advances, existing work largely overlooks the role of RLVR on synthetic data, particularly in the code domain. 
Since code naturally exhibits strong structural and verifiable properties, most prior approaches rely heavily on limited pools of real-world or curated datasets. 
This reliance constrains the scalability of RLVR and restricts its potential for further improving code reasoning and generation capabilities. 
In contrast, systematically leveraging synthetic data under RLVR remains underexplored, leaving a significant gap in current research.

\subsection{Synthetic Code Data Generation}

High-quality code data is essential for improving the programming capabilities of LLMs, but the high cost and limited scalability of manual annotation have motivated extensive research on synthetic data generation~\cite{villalobos2024position,zhao2025absolute,yue2025does}.
Most existing methods primarily focus on the pretraining and instruction fine-tuning stages, while largely ignoring the verifiability of code. 
These methods can be broadly categorized into two classes:
model-driven expansion methods, which iteratively rewrite or generate problems using LLMs (e.g., Evol-Instruct~\cite{luo2024wizardcoder}, Code Alpaca~\cite{codealpaca}, KodCode~\cite{xu-etal-2025-kodcode}, AutoCode~\cite{zhou2025autocode}, UniCode~\cite{zheng2025unicode}), 
and knowledge-based expansion methods, which synthesize data using limited external knowledge or structured signals (e.g., Package Instruct~\cite{huang2025opencoder}, Educational Instruct~\cite{huang2025opencoder}, OSS-Instruct~\cite{wei2024magicoder}).
Despite the effectiveness of RLVR as a post-training paradigm for code modeling, little work explores the use of synthetic data during the RL stage.
Furthermore, existing synthetic data largely samples near real-world data distributions, limiting diversity~\cite{xu-etal-2025-kodcode}, difficulty, and originality, and thus may provide insufficient learning signals for sustained RLVR optimization.

\section{Method}

In this section, we present a detailed overview of the \textit{\textbf{A}tomic \textbf{D}ecomposition and \textbf{R}ecombination} (ADR) framework (Figure~\ref{fig:main}). Specifically, ADR constructs a high-quality corpus of verifiable code data through element extraction, controllable element recombination, template-based problem synthesis, and execution-grounded validation. 
To improve the rigor and testing quality of the synthesized data, we introduce \textit{Adversarial Solution Space Refinement}. 
In addition, ADR employs \textit{Info-Guided Element Schema Optimization} to iteratively refine the element schema. 
Together, these components form a closed-loop framework that balances diversity, correctness, and difficulty. The implementation details can be found in Appendix~\ref{sec:adr_prompts}.

\subsection{Atomic Decomposition and Recombination}

ADR explicitly models code tasks as element compositions and explores the element space via controlled recombination. 
ADR decomposes generation into five stages: element extraction, controlled recombination, template-based problem synthesis, execution-grounded validation, and adversarial solution space refinement. 

\paragraph{Step 1: Element Extraction.}

We formalize the code problem space by defining task-specific element schemas and iteratively optimizing them. This stage consists of two primary phases: initial extraction and info-guided refinement.

\textit{1) Schema Definition and Extraction.}
A schema $\mathcal{S} = \{ e_1, \dots, e_n \}$ consists of $n$ elements definition, where each $e_i = (n_i, d_i, v_i)$ represents a semantic name, definition, and variation axis. 
We first leverage an LLM to generate an initial schema $\mathcal{S}^{(0)}$ based on task-specific characteristics. We then sample high-quality seed instances and decompose them into constituent elements according to $\mathcal{S}^{(0)}$.

\textit{2) Info-Guided Schema Optimization.}
To refine $\mathcal{S}^{(0)}$ without intensive manual effort, we introduce an automated optimization loop leveraging information-theoretic signals from the extracted elements.

\begin{itemize}[leftmargin=16pt, topsep=4pt]
    \item Probability Estimation: To compute information signals over textual elements, we first encode them using \texttt{all-MiniLM-L6-v2} and discretize into semantic clusters via K-Means. The probability $p(x_i)$ is estimated by the cluster frequency in the dataset.
    \item Global Diversity via Entropy ($H$): We compute $H(e_i) = -\sum_{j} p(c_j) \log p(c_j)$ for each element type, which guides split operations for over-concentrated clusters and merge operations for sparse ones.
    \item Logical Contribution via CMI ($I$): To quantify the marginal information a candidate element $e_i$ provides to the problem $q$ given the existing element $e_j$, we compute:
    \begin{equation}
        I(e_i; q \mid e_j) = \sum p(e_i, q, e_j)\log \frac{p(e_j)p(e_i, q, e_j)}{p(e_i, e_j)p(q, e_j)}
    \end{equation}
    which filters elements with negligible gain (remove) and prioritizes those that increase task complexity (add/redefine).
    \item Iterative Refinement: Guided by $\{H, I\}$, the LLM generates a sequence of refinement operations $\mathcal{O} \in \{\text{add, remove, split, merge, redefine}\}$ in a structured JSON format. These operations are automatically applied to update $\mathcal{S}^{(t+1)}$ until the average schema entropy $\bar{H}(\mathcal{S})$ converges or $t$ reaches $T_{max}$.
\end{itemize}

\paragraph{Step 2: Controlled Element Recombination.}

We address the risk of unsolvable combinations by anchoring generation around a core element $e_{\text{core}} \in \mathcal{S}$. 
The core element is selected by an LLM according to two criteria: (i) high information content and (ii) minimal coupling with other elements, maximizing its recombination flexibility.

Based on the core element $e_{\text{core}}^{(i)}$ and a small set of exemplar combinations $\{ C^{(j)} \}_{j=1}^3$, we prompt the LLM to generate a new combination $C_{\text{new}}$:

\begin{equation}
C_{\text{new}} \sim p_\theta^{\text{LLM}} \bigl( C_{\text{new}} \mid e_{core}^{(i)}, \{ C^{(j)} \}_{j=1}^3 \bigr).
\end{equation}

This method efficiently explores the element space, generating diverse yet semantically coherent problems without producing contradictory combinations.

\paragraph{Step 3: Template-Based Problem Synthesis.}

Given a generated element combination $D_{\text{new}}$, we produce a well-defined code problem using template-based synthesis, avoiding ambiguities common in free-form generation. 
Specifically, the LLM generates a problem $Q$ conditioned on the combination $C_{\text{new}}$ and a predefined template $T$, where $T$ specifies required fields (e.g., description, I/O format, constraints):

\begin{equation}
    Q \sim p_\theta^{\text{LLM}} ( Q \mid C_{\text{new}}, T ),
\end{equation}

\paragraph{Step 4: Execution-Grounded Validation.}

To ensure task validity and provide reliable feedback signals, we filter synthesized problems via execution, retaining only well-defined and solvable tasks.
Given a problem $Q$, the LLM generates a reference solution $\textit{sol}$ and a test case generator $G_{test}$:

\begin{equation}
(sol, G_{test}) \sim p_\theta^{\text{LLM}} ( sol, G_{test} \mid Q ).
\end{equation}

We execute $G_{test}$ to generate test cases $\mathcal{T} = { (x_i, y_i) }_{i=1}^N$, and validate the solution in an isolated sandbox, retaining only those with $\text{Valid}(Q)=1$:

\begin{equation}
    \text{Valid}(Q) =
    \begin{cases}
    1, & \text{if } \forall (x_i, y_i) \in \mathcal{T},\ sol(x_i) = y_i, \\
    0, & \text{otherwise}.
    \end{cases}
\end{equation}

\paragraph{Step 5: Adversarial Solution Space Refinement.}

To further enhance test coverage and robustness of synthetic data, we introduce an adversarial refinement stage.
Specifically, for a problem-solution pair, we first prompt the LLM to generate a set of \textit{near-miss solutions} $\mathcal{V} = \{v_1, v_2, \dots, v_k\}$, which are flawed by ignoring edge cases or making incorrect assumptions.
Then, we evaluate these solutions against the current test case $\mathcal{T}$ and compute the \textit{near-miss rate} $R(\mathcal{V}, \mathcal{T})$, defined as the proportion of flawed solutions that erroneously pass $\mathcal{T}$:
\begin{equation}
    R(\mathcal{V}, \mathcal{T}) = \frac{1}{|\mathcal{V}|} \sum_{v \in \mathcal{V}} \mathbb{I} \left[ \text{Fail}(v, \mathcal{T}) \right].
\end{equation}
To minimize $R$, we iteratively refine the test case generator $G_{test}$. Solutions in $\mathcal{V}$ that bypass the current tests are provided to prompt the LLM, which then synthesizes an updated $G_{test}$. This generator is then executed to produce new test cases. 
The process iterates until $R$ converges or reaches a predefined threshold. 

\section{Evaluation of Synthetic Data Quality}

To objectively evaluate the quality of synthetic code data, we first propose a multi-dimensional evaluation taxonomy. Based on this taxonomy, we compare several synthetic data baselines, and finally verify the effectiveness of ADR-synthesized data.

\subsection{Evaluation Taxonomy}

\label{sec:dataset_quality_metrics}

Synthetic code data quality is inherently multi-dimensional, requiring simultaneous consideration of novelty, challenge, coverage, and supervision reliability. We therefore design a four-dimensional taxonomy that captures originality, difficulty, diversity, and test quality.
Table~\ref{tab:metrics} provides the formal definitions of these metrics.
Detailed implementation details are provided in Appendix~\ref{sec:evaluation}.

\begin{table*}[t]
\centering
\renewcommand{\arraystretch}{1.3}
\caption{Evaluation taxonomy for synthetic data quality}
\label{tab:metrics}
\small
\begin{tabularx}{\textwidth}{l X p{5.0cm}}
\toprule
\textbf{Metric} & \textbf{Definition} & \textbf{Description} \\
\midrule
Originality & 
$|\mathcal{S}|^{-1} \sum_{x \in \mathcal{S}} \mathbb{I}\left( \max_{y \in \mathcal{R}} \cos(\phi(x), \phi(y)) < \tau \right)$ & 
Novelty relative to a reference dataset $\mathcal{R}$ with threshold $\tau=0.6$. \\
Difficulty & 
$1-|\mathcal{M}|^{-1} \sum_{m \in \mathcal{M}} \text{Perf}(m, \mathcal{S})$ & 
Task hardness computed by average performance of reference models $\mathcal{M}$. \\
Diversity & 
$1 - \sigma(d_i^{\text{nn}}) / \mu(d_i^{\text{nn}})$ & 
Uniformity of data distribution based on nearest-neighbor distances $d_i^{\text{nn}}$. \\
Test Quality & 
$|\mathcal{S}|^{-1} \sum_{x \in \mathcal{S}} c(x)$ & 
Test case diversity and boundary score $c(x)$ formulated by LLM-as-a-judge. \\
\bottomrule
\end{tabularx}
\vspace{-8pt}
\end{table*}

\subsection{Data Quality Evaluation Results}

\begin{wrapfigure}{r}{0.46\columnwidth}
    \centering
    \includegraphics[width=0.44\textwidth]{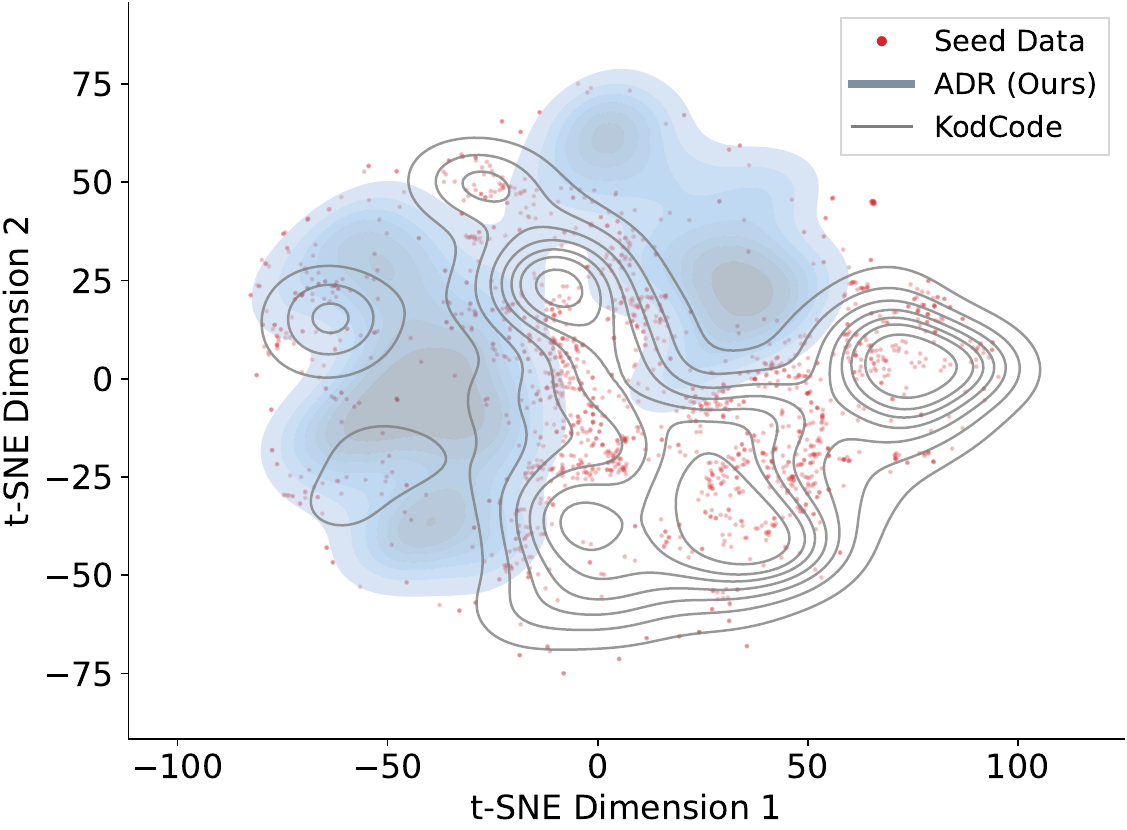}
    \caption{t-SNE visualization of data density coverage about ADR and KodCode data, both derived from the same seed data.}
    \label{fig:tsne_kde_density}
    \vspace{-5pt}
\end{wrapfigure}

\begin{wraptable}{r}{0.5\textwidth}
  \centering
  \vspace{-8pt}
  \caption{The data quality evaluation results.}
  \label{tab:analysis_task_quality}
  \small
  \begin{tabular}{lcccc}
    \toprule
    & Orig. & Diff. & Div. & T-Qual. \\
    \midrule
    KodCode & 1.78 & 17.92 & 72.75 & 29.91 \\
    Edu. Instr. & 6.04 & 20.14 & 46.17 & 37.82 \\
    \textbf{ADR} & \textbf{28.91} & \textbf{71.89} & \textbf{84.36} & \textbf{81.36} \\
    \bottomrule
  \end{tabular}
\end{wraptable}
Based on the evaluation framework, we find that \textbf{ADR benefits from element decomposition and recombination, leading to markedly improved synthetic data quality} across originality, difficulty, diversity, and test quality (Table~\ref{tab:analysis_task_quality}). 
Notably, ADR achieves an originality score of 28.91, significantly outperforming the strongest baseline, Educational Instruct with only 6.04.
In addition, Figure~\ref{fig:tsne_kde_density} compares t-SNE~\cite{maaten2008visualizing} visualization of data density coverage about ADR and KodCode synthesized from the same seed dataset (TACO~\cite{li2023taco}). 
ADR (blue region) exhibits a broader manifold, extending into long-tail regions beyond the high-density core, indicating more effective exploration. 
In contrast, KodCode~\cite{xu-etal-2025-kodcode} (gray contours) concentrates on localized areas with limited boundary coverage, reflecting a more conservative sampling strategy.
Although KodCode covers regions beyond ADR, 47.5\% of its unique samples  (KodCode-only) are simple function-level completions, while ADR focuses on more complex, instruction-style algorithmic tasks.
Further RL experiments (100 steps) following Section~\ref{sec:exp_alg_setup} show that ADR-only data leads to substantially larger performance gains on LCB-v5 (16.17$\rightarrow$\textbf{20.28}) than KodCode-only data (16.17$\rightarrow$17.89).

\subsection{Analysis of ADR components}

To validate the effectiveness of \textit{Info-Guided Element Schema Optimization (ESO)} in Step 1, we examine whether optimized element schemas improve synthesized data quality. 
Specifically, we randomly sample 100 instances from the TACO dataset as seed data. 
For each iteration, we generate 100 problems via controlled element recombination under the ASR paradigm. 
We then evaluate the diversity and validity rate of problems.
The evaluation results are in Table~\ref{tab:analysis_optimization}. 
As the optimization proceeds, we observe a consistent improvement in problem diversity across iterations. 
Moreover, the validity rate of the synthesized problems increases from 35.0\% to 43.0\%. 
These results demonstrate that iteratively optimizing element schemas using ESO effectively guides ADR in generating higher-quality synthetic data.

To evaluate the effectiveness of \textit{Execution-Grounded Validation}  in Step 4, we use LCB-v5 to determine whether the generated solutions pass the ground-truth test cases. 
Specifically, we randomly sampled 300 LCB-v5 instances; via only a single round of generation, we obtained 160 valid samples. Among these, the solutions achieved a 90.62\% pass rate on ground-truth test cases, demonstrating ADR's ability to produce reliably verifiable solutions.
\begin{wraptable}{r}{0.45\textwidth}
  \centering
  \vspace{-8pt}
  \caption{The evaluation results of synthetic problems in ESO iterations.}
  \label{tab:analysis_optimization}
  \small 
  \begin{tabular}{lcc}
    \toprule
    & Diversity & Validity \\
    \midrule
    ADR (iter1) & 85.54 & 35.0 \\
    ADR (iter2) & 85.62 & 32.0 \\
    \textbf{ADR (iter3)} & \textbf{86.70} & \textbf{43.0} \\
    \bottomrule
  \end{tabular}
\end{wraptable}

To evaluate the effectiveness of \textit{Adversarial Solution Space Refinement (ASSR)} in Step 5, we focus on both the quantity and quality of test cases, as defined in Section~\ref{sec:dataset_quality_metrics}. 
Applying ASSR to 5K ADR-synthesized tasks increases the average number of test cases from 14.75 to 34.78 (+135.8\%), and improves test quality from 72.91 to 81.36 (+11.6\%).
These gains indicate that ASSR substantially enhances the effectiveness of generated test cases.

\section{RLVR Experiments}

\subsection{Experimental Setup}
\label{sec:exp_alg_setup}

\begin{table*}[t]
  \centering
  \caption{Pass@1 (\%) performance comparison on algorithmic tasks across multiple benchmarks and representative base models. Results show that 
  prior synthetic data methods often fail to outperform original data training, while ADR consistently achieves better overall performance across models. $\dagger$ denotes a statistically significant improvement over baselines ($p < 0.001$, McNemar’s test).}
    \begin{tabular}{l>{\centering\arraybackslash}p{2.5cm}>{\centering\arraybackslash}p{2cm}>{\centering\arraybackslash}p{2cm}}
    \toprule
          & \textbf{LCB-v5} & \textbf{LCB-v6} & \textbf{Average} \\
    \midrule
    Qwen2.5-Coder-7B-Instruct & 16.17  & 20.21  & 18.19  \\
    + TACO \tiny{\textit{Real}} & 22.60  & \underline{23.86}  & \underline{23.23}  \\
    + Educational Instruct \tiny{\textit{Synthetic}} & 19.61  & 21.71  & 20.66  \\
    + KodCode \tiny{\textit{Synthetic}} & \underline{22.75}  & 23.57  & 23.16  \\
    \rowcolor{blue!5}
    \textbf{+ ADR} (ours) \tiny{\textit{Synthetic}} & \textbf{25.37}$^{\dagger}$ & \textbf{26.14}$^{\dagger}$ & \textbf{25.76}$^{\dagger}$ \\
    \midrule
    Llama-3.1-8B-Instruct & 9.36  & 15.71  & 12.54  \\
    + TACO \tiny{\textit{Real}} & 10.25  & 14.21  & 12.23  \\
    + Educational Instruct \tiny{\textit{Synthetic}} & 8.01  & 14.71  & 11.36  \\
    + KodCode \tiny{\textit{Synthetic}} & \underline{12.20} & \underline{17.93}  & \underline{15.06}  \\
    \rowcolor{blue!5}
    \textbf{+ ADR} (ours) \tiny{\textit{Synthetic}} & \textbf{16.84}$^{\dagger}$ & \textbf{23.00}$^{\dagger}$ & \textbf{19.92}$^{\dagger}$ \\
    \midrule
    Qwen3-8B & 22.53  & 21.21  & 21.87  \\
    + TACO \tiny{\textit{Real}} & \underline{34.81}  & \underline{27.43}  & \underline{31.12}  \\
    + Educational Instruct \tiny{\textit{Synthetic}} & 25.15  & 23.50  & 24.32  \\
    + KodCode \tiny{\textit{Synthetic}} & 26.27  & 24.57  & 25.42  \\
    \rowcolor{blue!5}
    \textbf{+ ADR} (ours) \tiny{\textit{Synthetic}} & \textbf{35.85}$^{\dagger}$ & \textbf{31.43}$^{\dagger}$ & \textbf{33.64}$^{\dagger}$ \\
    \bottomrule
    \end{tabular}%
    \vspace{-8pt}
  \label{tab:main_exp_algorithm}%
\end{table*}%

\textbf{Baselines.}
We compare our ADR-based model against several widely used baselines. These include synthetic-data baselines, KodCode~\cite{xu-etal-2025-kodcode} (algorithm, data structure, and package subset) and Educational Instruct~\cite{huang2025opencoder}, which provide verifiable signals to support RL training, as well as a real-data baseline, TACO~\cite{li2023taco}, which is commonly adopted as seed data for synthetic data methods.

To comprehensively demonstrate the effectiveness of the ADR synthesis paradigm, we conduct comparative experiments on multiple representative base models (\textit{Qwen2.5-Coder-7B-Instruct}~\cite{hui2024qwen2}, \textit{Llama-3.1-8B-Instruct}~\cite{grattafiori2024llama}, and \textit{Qwen3-8B} (Non-thinking)~\cite{yang2025qwen3}) and across various code domains (algorithms, tool usage, and data science).

\textbf{RL Setup.}
For algorithm tasks, we randomly sample 5,000 examples from each baseline for training. 
For ADR, we select 1,710 verified TACO problems with difficulty above medium as seed data and synthesize 5,000 training data using DeepSeek-V3.2~\cite{liu2025deepseek}.
For tool usage and data science tasks, we randomly select 5,000 problems from Package Instruct~\cite{huang2025opencoder} as seed data and synthesize 2,000 training data using DeepSeek-V3.2. 
Since Package Instruct is designed for SFT and lacks verifiable signals, it cannot be directly used for RL training. 
More data details can be found in Appendix~\ref{sec:adr_data}.

For training, we perform the GRPO~\cite{shao2024deepseekmath} algorithm for 10 epochs, with a global batch size of 128, a mini batch size of 32, 8 rollouts per question, learning rate of 1e-6 and max response length of 8192.

\textbf{Evaluation.}
For algorithm tasks, we evaluate models on LiveCodeBench~\cite{jainlivecodebench}, which measures the ability to generate competitive programming solutions. We use the widely adopted v5 (2410–2501) and v6 (2501–2504) subsets, consisting of 167 and 175 problems, respectively.
For the tool usage tasks, we evaluate model performance on the widely used BigCodeBench~\cite{zhuobigcodebench}, which measures the ability to follow complex, real-world instructions. For data science tasks, we use DS-1000~\cite{lai2023ds} to assess the model’s capability to generate data-processing code using data science libraries.
We sample 8 times per problem, with max output length of 32768, temperature of 0.6, and top\_p of 0.95.

\subsection{Overall Results}

Table~\ref{tab:main_exp_algorithm} and Table~\ref{tab:main_exp_tool_calling_data_science} show the RL performance comparison between ADR-synthesized data and multiple baselines across algorithmic, tool usage, and data science tasks.

\begin{wrapfigure}{r}{0.46\columnwidth}
    \centering
    \includegraphics[width=0.46\columnwidth]{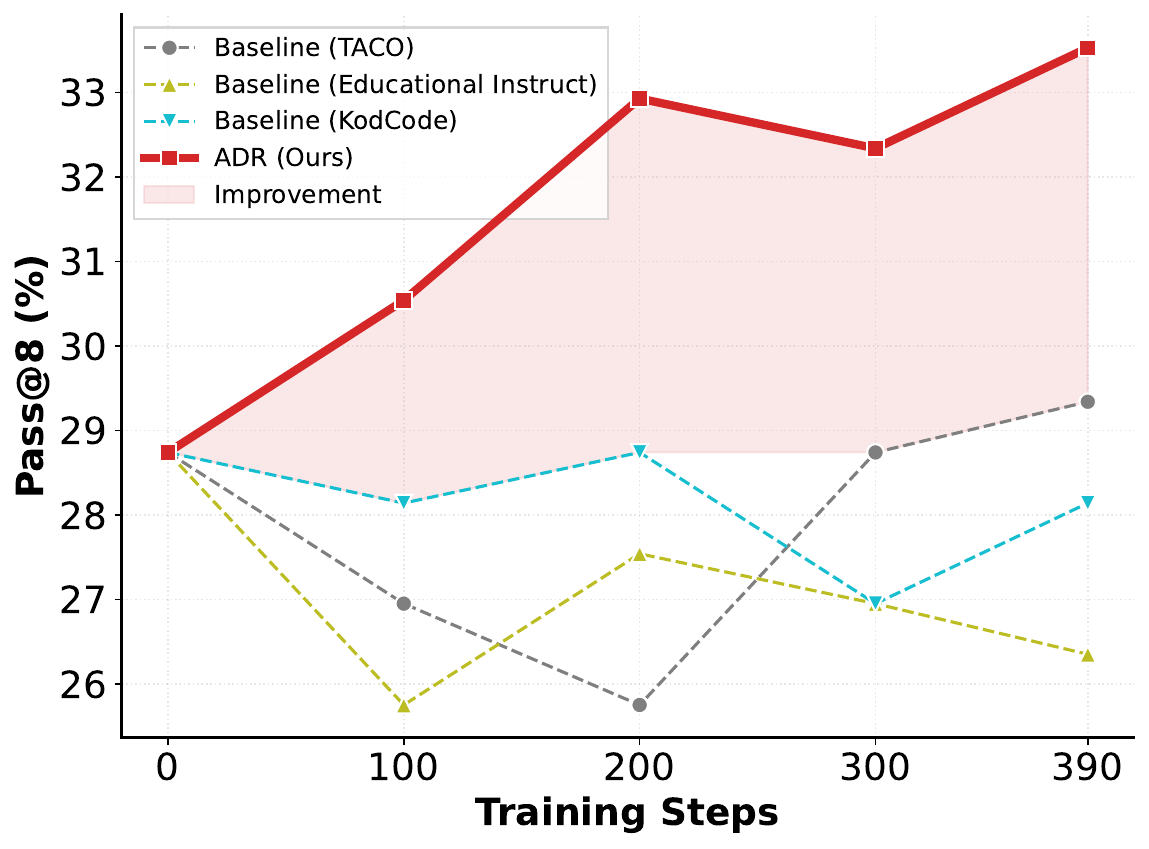}
    \caption{Pass@8 (\%) performance on LCB-v5 for Qwen2.5-Coder-7B-Instruct.}
    \label{fig:pass_at_8}
    \vspace{-12pt}
\end{wrapfigure}
\textbf{Previous synthetic data methods, constrained by heuristic expansions of real-world data, fail to surpass original data performance.}
For example, on LCB-v5 and LCB-v6, Educational Instruct achieves an average of 20.66\% on \textit{Qwen2.5-Coder-7B-Instruct}, substantially underperforming TACO (23.23\%), while the strongest baseline, KodCode, reaches 23.16\%, only on par with TACO. 
Moreover, on Qwen3-8B, KodCode and Educational Instruct exhibit relative performance drops of 5.70\% and 6.80\% compared to TACO. 
On Llama-3.1-8B-Instruct, TACO and Educational Instruct suffer from reward saturation during training, leading to large gradients and unstable optimization, which results in performance degradation even below the base model. 
These results suggest that current data synthesis methods remain ineffective, as they fail to truly explore the atomic components of problems, limiting the utility of the synthesized data.

\textbf{Benefiting from element decomposition and recombination, ADR-synthesized data achieves superior overall performance.}
Our ADR-based model consistently outperforms all baselines across various benchmarks, with particularly strong gains over the real-data baseline. 
For example, on LCB-v5, ADR achieves 25.37\% (+9.20\%) on \textit{Qwen2.5-Coder-7B-Instruct}, while the best synthetic-data baseline reaches 22.75\% (+6.58\%), and the real-data baseline achieves 22.60\% (+6.43\%). 
Similarly, on LCB-v6, ADR attains 26.14\%, surpassing the strongest baseline at 23.86\%. 
Overall, ADR achieves an overall 25.76\%, substantially higher than the best baseline’s 23.23\%, demonstrating that the ADR synthesis paradigm can efficiently explore the element space and generate high-quality training data from limited seed examples for RL optimization.

\textbf{ADR-synthesized data expands the model’s intrinsic reasoning capacity.}
As shown in Figure~\ref{fig:pass_at_8}, with increased sampling, the ADR-based model exhibits substantially larger improvements (28.74\% to 33.53\%, +4.79\%), while the strongest baseline TACO achieves only +0.60\%. This demonstrates that ADR effectively synthesizes data targeting the model’s boundary capabilities, consistent with the perspectives on RL effectiveness discussed in prior work~\cite{zhang2025interplay}.

\textbf{ADR demonstrates robust generalizability, yielding consistent performance gains across multiple base models.}
For example, on \textit{Qwen2.5-Coder-7B-Instruct}, ADR yields an improvement of 7.57\%. The gains remain significant on \textit{Llama-3.1-8B-Instruct} and \textit{Qwen3-8B}, reaching 7.38\% and 11.77\%, respectively, and exceeding the best baseline improvements of 2.52\% and 9.25\%. 
These results indicate that ADR produces broadly effective synthetic data and generalizes well across different base model architectures.

\begin{wraptable}{r}{0.5\textwidth}
  \centering
  \caption{Pass@1 (\%) performance on tool usage and data science tasks. $\dagger$ denotes a statistically significant improvement over KodCode ($p < 0.001$, McNemar’s test).}
  \label{tab:main_exp_tool_calling_data_science}
  \small
  \setlength{\tabcolsep}{4pt}
  \begin{tabular}{lcc}
    \toprule
    & \textbf{Tool Usage} & \textbf{Data Science} \\
    \cmidrule(lr){2-2} \cmidrule(lr){3-3}
    & BigCodeBench & DS-1000 \\
    \midrule
    Qwen2.5-7B-Ins & 38.30 & 36.28 \\
    + KodCode      & 41.27 & 39.05 \\
    \rowcolor{blue!5}
    + \textbf{ADR (ours)} & \textbf{41.67}$^{\dagger}$ & \textbf{42.44}$^{\dagger}$ \\
    \bottomrule
  \end{tabular}
\end{wraptable}

\begin{table*}[t]
    \centering
    \caption{A comparative case study of problem synthesis paradigms. While heuristic expansion-based tasks maintain the seed's core elements (e.g., Hamming distance, prefix sums) and merely alter constraints, the ADR-based task undergoes a significant structural transformation.}
    \vspace{-5pt}
    \setlength{\fboxsep}{2pt}
    \resizebox{\linewidth}{!}{
    \begin{tabular}{p{0.12\linewidth} p{0.60\linewidth} p{0.28\linewidth}}
    \toprule
    \textbf{Task Type} & \textbf{Problem Description} & \textbf{Core Elements} \\
    \midrule
    Seed Task & \underline{Genos} needs your help ... The Hamming distance between two strings s and t ... Given two \textbf{binary strings} a and b, find the sum of the \textbf{Hamming distances} between a and all \textbf{contiguous substrings} of b of length $|a|$ ... & 
    \parbox[t]{0.28\linewidth}{
    \raggedright
    \colorbox{gray!15}{Binary strings} \par\vspace{2pt}
    \colorbox{gray!15}{Hamming distance} \par\vspace{2pt}
    \colorbox{gray!15}{Sliding window+Sum.} \par\vspace{2pt}
    \colorbox{gray!15}{Prefix sums}
    } \\
    \midrule
    Heuristic expansion-based Task & Saitama has given \underline{Genos} another intriguing challenge ... Consider a \textbf{binary string} b of length $|b|$ ... help \underline{Genos} find the \textbf{minimum Hamming distance} required to transform b into either of the valid \textbf{alternating patterns} "010101..." or "101010..." ... & 
    \parbox[t]{0.28\linewidth}{
    \raggedright
    \colorbox{gray!15}{Binary strings} \par\vspace{2pt}
    \colorbox{gray!15}{Hamming distance} \par\vspace{2pt}
    \colorbox{green!15}{Fixed}~\colorbox{gray!15}{window} \par\vspace{2pt}
    \colorbox{gray!15}{Prefix sums}
    } \\
    \midrule
    ADR-based Task & A textile factory uses an automated machine to inspect fabric rolls for defects ... For each roll `i`, it computes a \textbf{dissimilarity score} `a[i]` against a reference template ... The machine examines every \textbf{contiguous segment} ... identifies the \textbf{maximum dissimilarity score} within that window ... & 
    \parbox[t]{0.28\linewidth}{
    \raggedright
    \colorbox{green!15}{Integer array} \par\vspace{2pt}
    \colorbox{green!15}{Dissimilarity score} \par\vspace{2pt}
    \colorbox{gray!15}{Sliding window}+\colorbox{green!15}{Max} \par\vspace{2pt}
    \colorbox{green!15}{Monotonic queue}
    } \\
    \bottomrule
    \end{tabular}
    }
    \label{tab:case_study}
    \vspace{-6pt}
\end{table*}

\textbf{ADR enables cross-domain generalization.}
The ADR-based model yields consistent gains across task categories, improving performance on tool usage (41.67\%, +3.37\%) and data science (42.44\%, +6.16\%) over KodCode.
These results demonstrate that ADR generalizes beyond algorithmic programming and remains effective on more diverse code-centric tasks.
While most prior work on code RL focuses primarily on algorithmic problems, our results indicate that ADR provides a practical path to extend RL training to broader domains, promoting more general-purpose coding capabilities.

\subsection{Case Studies}

To further investigate ADR’s capability to generate genuinely novel problems compared to heuristic expansion-based methods, we conduct a comparative case study (Table~\ref{tab:case_study}) using the same seed tasks for both KodCode and ADR.
Heuristic expansion-based methods maintain a high degree of structural similarity to the seed task.
Although the objective may change (e.g., from computing a sum to minimizing Hamming distance), it remains within the same data types and core operations.
Such variations are largely incremental, involving minor adjustments to the problem's constraints rather than its underlying logic.
In contrast, \textbf{ADR enables structural innovation through controlled decomposition and recombination}. 
It frequently introduces new data structures (e.g., binary strings to integer arrays) and shifts algorithmic paradigms (e.g., prefix sums to monotonic queues). 
By exploring a broader structured design space, ADR moves beyond local modifications and produces tasks with fundamentally new algorithmic requirements.

\subsection{Detailed Analysis}

\begin{figure*}[!t]
\centering
\begin{subfigure}{0.32\columnwidth}
    \centering
    \small
    \includegraphics[width=\textwidth]{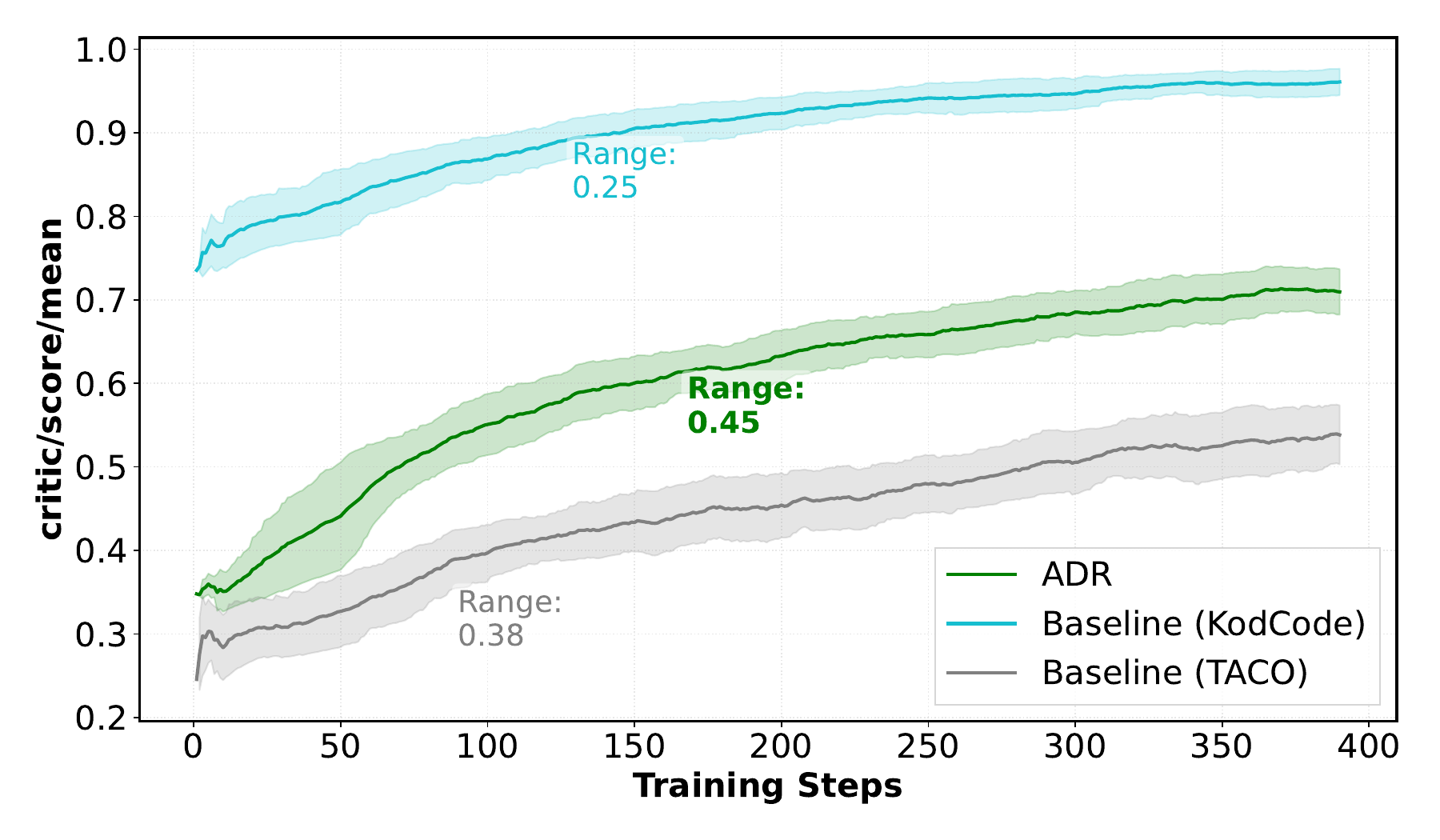}
    \caption{Reward curve.}
    \label{fig:critic_score_mean}
\end{subfigure}
\begin{subfigure}{0.32\columnwidth}
    \centering
    \small
    \includegraphics[width=\textwidth]{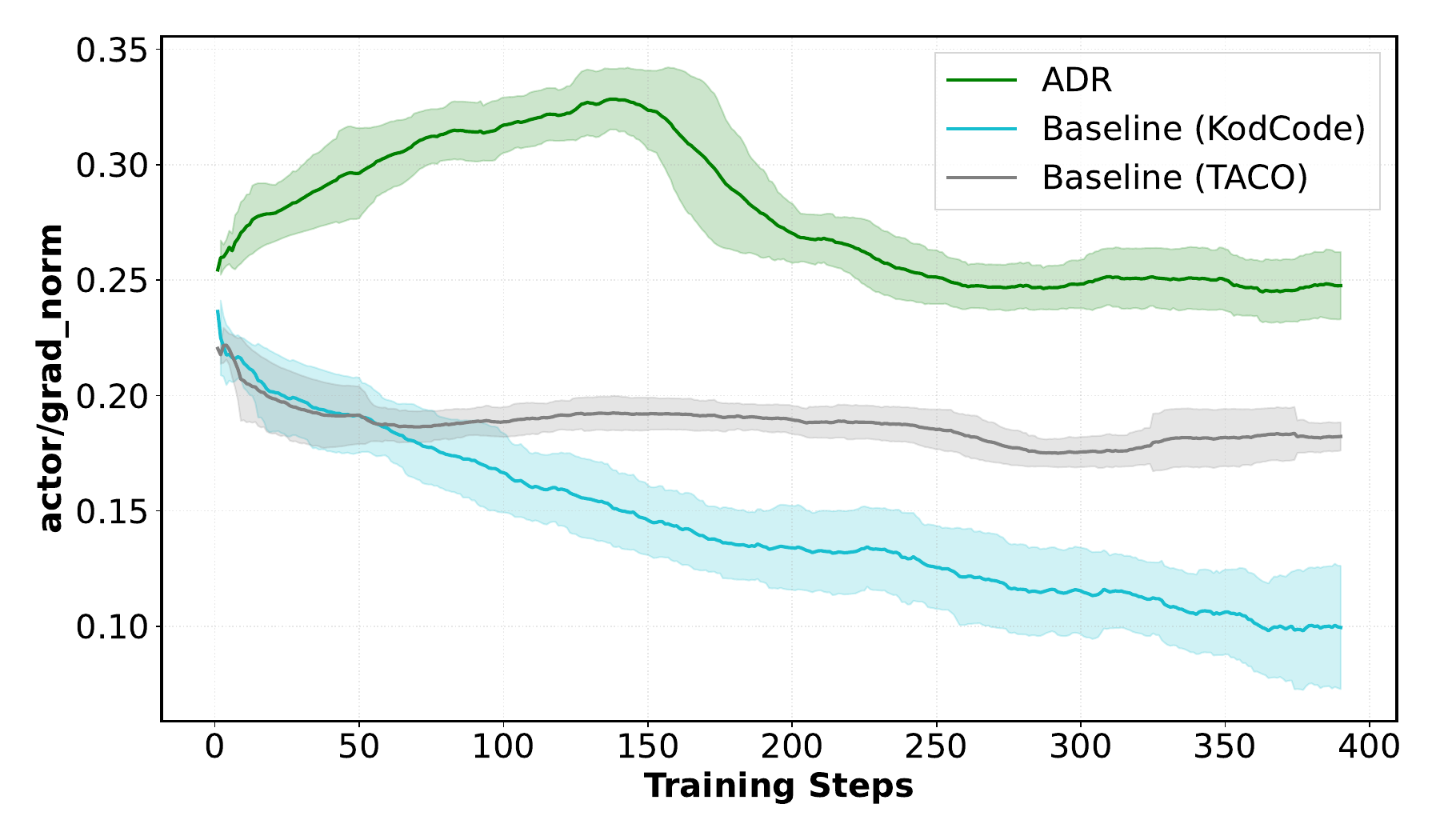}
    \caption{Actor gradient norm.}
    \label{fig:actor_grad_norm}
\end{subfigure}
\hfill
\begin{subfigure}{0.32\columnwidth}
    \centering
    \small
    \includegraphics[width=\textwidth]{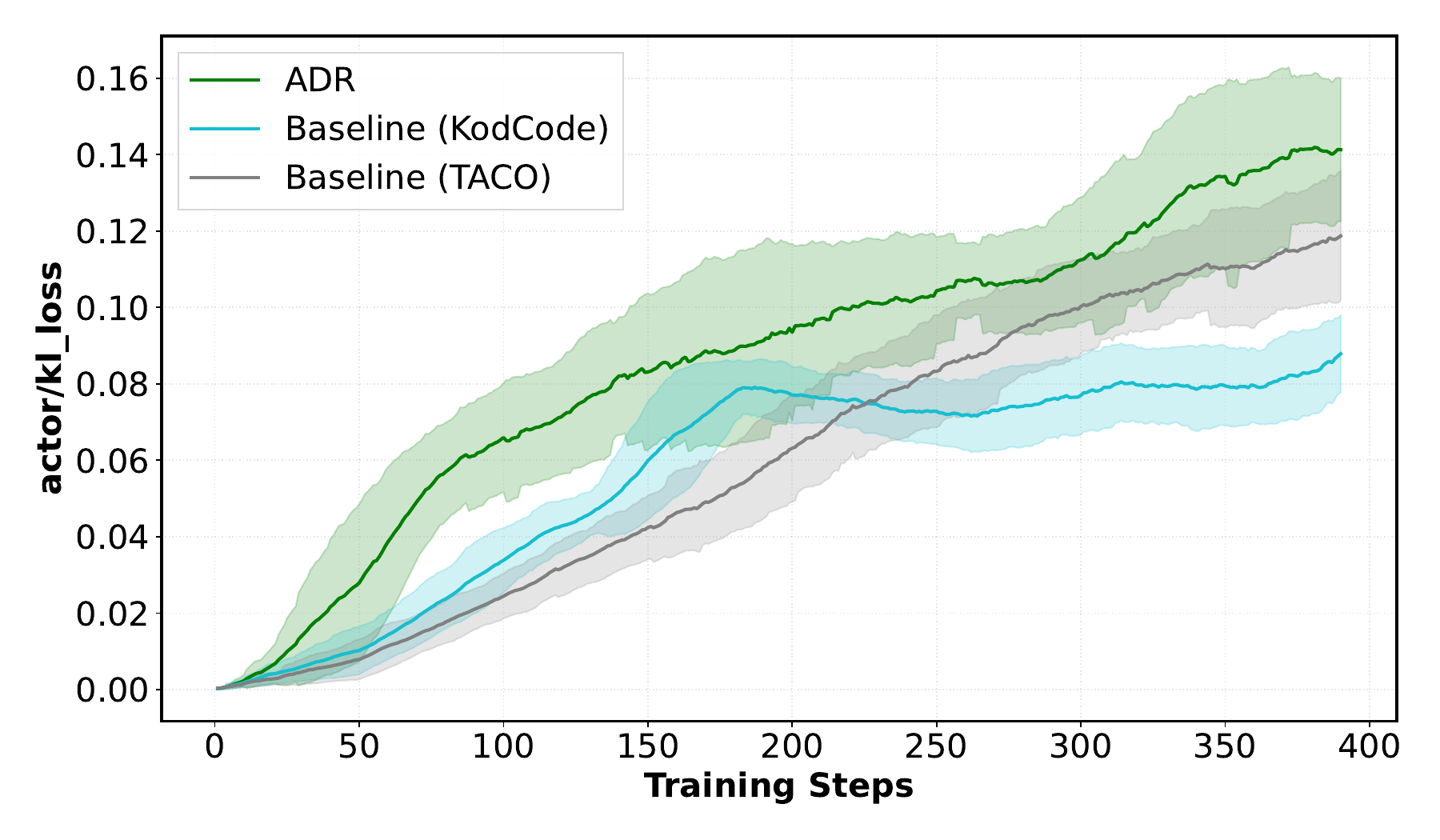}
    \caption{Actor KL loss.}
    \label{fig:actor_kl_loss}
\end{subfigure}
\caption{RL training dynamics of ADR and baseline datasets based on Qwen2.5-Coder-7B-Instruct.}
\vspace{-8pt}
\label{fig:analysis}
\end{figure*}

\textbf{ADR enhances the optimization potential and maintains training updates throughout the RL process.}
As shown in Figure~\ref{fig:critic_score_mean}, while the KodCode baseline exhibits higher initial performance, its cumulative improvement remains relatively limited ($\Delta=0.25$). 
In contrast, ADR achieves a larger gain ($\Delta=0.45$), suggesting a more extensible optimization landscape that allows the model to bridge the gap from a lower baseline to high performance. 
This trend is further supported by the actor gradient norm in Figure~\ref{fig:actor_grad_norm}. 
Unlike the baselines where gradient norms decay in later stages, ADR maintains a stable plateau (approx. 0.25). 
This persistence indicates that ADR provides a steady stream of informative signals, even in the later stages of training, which helps mitigate premature convergence and supports continuous performance growth.

\textbf{ADR promotes deeper policy exploration.}
As in Figure~\ref{fig:actor_kl_loss}, the Actor KL loss trajectory demonstrates that ADR induces the strongest and smoothest linear growth, converging at approximately 0.14, considerably higher than the 0.08 observed for KodCode. 
A steady increase in KL divergence reflects the extent of policy evolution relative to the initial SFT model. 
Baseline data, with lower information density, leads the model to rapidly settle into local optima, causing policy drift to stagnate. 
In contrast, ADR’s richer logical structure and more discriminative reward signals guide the model beyond its original probability distribution, enabling substantive behavioral evolution.

\section{Conclusion}

In this work, we introduced ADR, a framework that overcomes the limitations of traditional data synthesis by decomposing code tasks into atomic elements and recombining them. By moving beyond heuristic seed expansions, ADR generates genuinely novel, challenging, and verifiably correct tasks that push the boundaries of LLM performance.
Our evaluations show that ADR significantly enhances data diversity and quality, leading to superior RLVR performance across algorithmic, tool usage, and data science. 
Ultimately, ADR provides a scalable paradigm for synthesizing the high-quality code data necessary to train the next generation of code LLMs.

\section*{Limitations}

Although ADR is designed as a fully automated framework capable of adapting to diverse code tasks, its current evaluation is limited to specific benchmarks and model scales. 
In the future, we plan to scale RLVR training to larger foundation models and multilingual environments to fully verify data robustness. Additionally, we will extend ADR from single-turn code generation to broader code agent scenarios, such as automated software engineering and multi-turn autonomous problem-solving.

\bibliographystyle{unsrt}
\bibliography{reference}


\newpage
\appendix
\section{Additional Experimental Setups}

\subsection{Evaluation of Synthetic Data Quality}
\label{sec:evaluation}

\begin{enumerate}
    \item Originality: We select the \texttt{PrimeIntellect/verifiable-coding-problems} as the reference dataset $\mathcal{R}$, which contains 144,169 problems spanning diverse sources, including Apps~\cite{hendrycks2measuring}, CodeContests~\cite{li2022competition}, Codeforces~\cite{jur1cek2023codeforces}, and TACO~\cite{li2023taco}. We obtain data representations using the \texttt{all-MiniLM-L6-v2} embedding model and compute cosine similarity.
    \item Difficulty: We choose \texttt{Qwen/Qwen3-4B}, \texttt{Qwen/Qwen3-8B}, and \texttt{Qwen/Qwen3-14B}~\cite{yang2025qwen3} as the representative model set $\mathcal{M}$. We choose the non-thinking mode.
    \item Diversity: We obtain data representations using the \texttt{all-MiniLM-L6-v2} embedding model and compute the Euclidean distance between each pair of problems. For each problem, we then identify its nearest neighbor in the embedding space and record the corresponding nearest-neighbor distance. We compute the coefficient of variation (CV) of all nearest-neighbor distances, defined as the standard deviation divided by the mean. A smaller CV indicates that inter-point distances are more uniform, corresponding to a more even distribution. The final score is defined as (1 - CV), where higher values indicate a smaller variation and a more uniformly distributed problem set.
    \item Test Quality: We evaluate test case quality using an LLM-as-a-Judge framework, considering both test case diversity and edge coverage. The corresponding prompt is shown in Figure~\ref{fig:eval_test_quality}.
\end{enumerate}

\subsection{ADR-synthesized Data in RLVR Experiments}
\label{sec:adr_data}

For algorithm tasks, we follow the ADR paradigm and synthesize 5,000 training data using DeepSeek-V3.2~\cite{liu2025deepseek}. Specifically, we select 1,710 verified TACO~\cite{li2023taco} problems with difficulty above the medium level (MEDIUM and MEDIUM\_HARD) in \texttt{PrimeIntellect/verifiable-coding-problems} as seed data. We verify the problems in an isolated sandbox.
Then, we perform 8-times controlled recombination for each seed problem. 
Finally, we apply execution-grounded validation to filter and retain 5,000 valid data.

For tool usage and data science tasks, we follow the ADR paradigm and both synthesize 2,000 training data using DeepSeek-V3.2~\cite{liu2025deepseek}. 
Specifically, we randomly sample 5,000 examples from Package Instruct~\cite{huang2025opencoder} for both. 
For tool-use tasks, we filter tools to retain those aligned with the BigCodeBench task taxonomy, while for data science tasks, we retain data science libraries consistent with the DS-1000 task taxonomy.
Then, we perform 1-time controlled recombination for each seed problem and filter 2,000 valid data.

\section{ADR Prompt Templates}
\label{sec:adr_prompts}

For the details of ADR steps across different tasks, Figures~\ref{fig:adr_prompt_algorithmic_step_1} to \ref{fig:adr_prompt_algorithmic_step_4} show prompt templates for algorithmic tasks, Figures~\ref{fig:adr_prompt_tool_step_1} to \ref{fig:adr_prompt_tool_step_4} for tool-use tasks, and Figures~\ref{fig:adr_prompt_data_step_1} to \ref{fig:adr_prompt_data_step_4} for data science tasks.

For Info-Guided Element Schema Optimization, the corresponding prompt template is shown in Figures~\ref{fig:adr_optimization_1} and \ref{fig:adr_optimization_2}.
For Adversarial Solution Space Refinement, the corresponding prompt template is shown in Figures~\ref{fig:adr_adversarial_step_1} and \ref{fig:adr_adversarial_step_2}.

\begin{tcolorbox}[breakable, width=1\textwidth, fontupper=\small, colback=blue!2, boxrule=0.9pt] 
\begin{lstlisting}[breaklines=true, basicstyle=\ttfamily\small]
You are an expert in software testing, program analysis, and algorithm evaluation.

## Task Description

You are given a {task_type} problem.
Your task is to **generate evaluation code** that computes a **test case coverage score** in the range **[0, 1]**, representing how well the test cases cover the problem's logic and edge cases.
The generated code will be used to **evaluate the quality of test cases**, not to solve the problem itself.

## Test Case Formats

The test cases will follow **one of the two formats** below:

### 1. `stdin_stdout`

```json
{{
    "input": inputs,      // List of stdin strings
    "output": outputs,    // List of stdout strings
    "fn_name": null,
    "type": "stdin_stdout"
}}
```

### 2. `function_call`

```json
{{
    "input": inputs,      // List of argument lists
    "output": outputs,    // List of return values
    "fn_name": "function_name",
    "type": "function_call"
}}
```

## Evaluation Dimensions

The evaluator must compute **exactly three sub-scores**, each normalized to **[0, 1]**:

### 1. Test Case Diversity Score (weight = 0.5)

Measures how **diverse and non-redundant** the test cases are:

* Variety of input sizes
* Variety of value distributions
* Structural diversity of arguments or stdin patterns
* Low duplication or trivial repetition

### 2. Edge Case Coverage Score (weight = 0.5)

Measures whether the test cases include **common edge cases**, such as:

* Empty or minimal inputs
* Maximum-sized inputs (if inferable)
* Boundary values (e.g., 0, -1, 1, max-int-like values)
* Special structural cases (single element, all equal, extreme imbalance)

## Mandatory Evaluation Code Template

You **must generate self-contained code strictly following the template below**.
You may add helper functions, but **must not change the function signatures**.

```python
def evaluate_testcase_coverage(testcases: dict) -> float:
    """
    Evaluate the coverage quality of algorithm test cases.

    Scoring dimensions (fixed):
    - Test case diversity (0.5)
    - Edge case coverage (0.5)

    Args:
        testcases (dict): Test cases in either `stdin_stdout`
                          or `function_call` format.

    Returns:
        float: Coverage score in range [0.0, 1.0]
    """

    # ----------------------------
    # 1. Basic validation
    # ----------------------------
    if not testcases or "input" not in testcases or "output" not in testcases:
        return 0.0

    inputs = testcases.get("input", [])
    outputs = testcases.get("output", [])
    tc_type = testcases.get("type", None)

    if not inputs or len(inputs) != len(outputs):
        return 0.0

    # ----------------------------
    # 2. Test case diversity score
    # ----------------------------
    diversity_score = 0.0

    # ----------------------------
    # 3. Edge case coverage score
    # ----------------------------
    edge_case_score = 0.0

    # ----------------------------
    # 4. Fixed-weight aggregation
    # ----------------------------
    final_score = (
        0.5 * diversity_score +
        0.5 * edge_case_score
    )

    return max(0.0, min(1.0, float(final_score)))
```

## Algorithm Problem
{problem}
\end{lstlisting}
\end{tcolorbox}

\captionof{figure}{Prompt Template for Test Quality Metric.}
\label{fig:eval_test_quality}

\begin{tcolorbox}[breakable, width=1\textwidth, fontupper=\small, colback=blue!2, boxrule=0.9pt] 
\begin{lstlisting}[breaklines=true, basicstyle=\ttfamily\small]
Please analyze the following algorithm problem according to the guidelines below. The output should follow the format below (do not add any other notes or explanations):
<answer>
Core Algorithm Idea:
Story Background:
Strategy Diversity:
Difficulty Level:
</answer>

### Guidelines
- Core Algorithm Idea:
    - Extract the essential algorithmic principle required to solve the problem.
    - Follow these rules:
        - Identify the **primary computational technique** (e.g., greedy strategy with a specific invariant, DP with defined states and transitions, graph traversal with constraints, combinatorial search, optimization structure).
        - Focus on the **abstract reasoning pattern**, not the implementation details.
        - Ensure the idea is **specific enough** to distinguish it from other techniques, but **general enough** to be reused in new problems.
        - Highlight key structural properties (optimal substructure, monotonicity, connectivity, constraint types, state formulation, etc.)
- Story Background:
    - Describe a simple and generic narrative theme that provides motivation or flavor for the problem in one abstract, conceptual sentence.
    - Follow these rules:
        - Be conceptual rather than operational (e.g., "managing resources", "tracking evolving states", "navigating a structure").
        - Avoid concrete rules, numeric details, domain-specific mechanics, or anything implying algorithmic constraints. 
        - Not depend on specific data types, input formats, or procedures - those belong to other elements. The background should be flexible enough to pair with many different algorithmic cores without creating conflicts.
- Strategy Diversity:
    - List the legitimate algorithmic approaches that could solve the problem **in principle**, and explain why.
    - Follow these rules:
        - Cover the full spectrum of viable strategies:
            - Greedy (local optimality conditions)
            - Dynamic Programming (state decomposition)
            - Graph search (DFS/BFS/backtracking)
            - Data-structure-driven optimization (segment tree, union-find, Fenwick tree, etc.)
            - Approximation/heuristics (if NP-hard structures are implied)
        - Explain the **structural justification** (e.g., overlapping subproblems, convexity, combinatorial explosion).
        - These should be **strategies that could plausibly apply**, not necessarily the optimal one.
- Difficulty Level:
    - Classify the problem's difficulty based on conceptual and implementation challenges.
    - Follow these rules:
        - Use the scale: **Beginner / Intermediate / Advanced**.
        - Consider:
            - Core concepts required
            - Input size limits
            - Edge-case density
            - Data structure sophistication
            - Theoretical complexity (polynomial vs NP-hard)
        - Justify the classification clearly.

### Algorithm Problem
{problem}

### Solution of the Algorithm Problem
```python
{solution}
```

### Analysis
\end{lstlisting}
\end{tcolorbox}

\captionof{figure}{Prompt Template for \textit{Step 1: Element Extraction} in ADR (algorithmic task).}
\label{fig:adr_prompt_algorithmic_step_1}

\begin{tcolorbox}[breakable, width=1\textwidth, fontupper=\small, colback=blue!2, boxrule=0.9pt] 
\begin{lstlisting}[breaklines=true, basicstyle=\ttfamily\small]
You are an expert in algorithmic problem design. Your task is to analyze the given Story Background and use the three provided combinations as inspiration, then construct one new and coherent set of four elements (Core Algorithm Idea, Story Background, Strategy Diversity, Difficulty Level).
The output should follow the format below (do not add any other notes or explanations):
<answer>
Core Algorithm Idea:
Story Background:
Strategy Diversity:
Difficulty Level:
</answer>

Requirements:
1. Learn the structural patterns, not the content
    - Extract from the examples their level of detail, reasoning style, and the way elements relate to each other. Do not reuse or merge their specific ideas.
2. Preserve internal coherence
    - Ensure the four generated elements naturally support each other:
        - The Story Background should organically introduce the constraints that motivate the Core Algorithm Idea.
        - The Strategy Diversity must correspond to the algorithmic structure implied by the Core Algorithm Idea.
        - The Difficulty Level should reflect the conceptual and technical depth of the chosen algorithmic direction.
3. Maintain originality and avoid conflicts
    - Your output must be a fully new construction - no copying from examples - and must avoid internal contradictions in constraints, methods, or complexity assumptions.
4. Ensure problem-level richness
    - The new combination should have enough structure and complexity to support a meaningful algorithm problem, with:
        - Non-trivial decisions or constraints
        - Multiple plausible solution approaches
        - Clear reasons for the difficulty classification

### Story Background
{story_background}

### Combinations Reference
{combinations_1}

{combinations_2}

{combinations_3}

### New Elements Combination
\end{lstlisting}
\end{tcolorbox}

\captionof{figure}{Prompt Template for \textit{Step 2: Controlled Recombination} in ADR (algorithmic task).}
\label{fig:adr_prompt_algorithmic_step_2}

\begin{tcolorbox}[breakable, width=1\textwidth, fontupper=\small, colback=blue!2, boxrule=0.9pt] 
\begin{lstlisting}[breaklines=true, basicstyle=\ttfamily\small]
You are an expert problem setter and algorithmist. Using only the provided Problem Framework, generate a single, self-contained, original, and challenging algorithm problem suitable for programming contests or practice platforms. Follow these rules carefully:

1. **Framework fidelity with creative flexibility**: The problem must strictly respect the content, constraints, and core setup defined in the Problem Framework. You may introduce new scenarios, examples, or problem twists only within the limits of the framework, in order to make the problem original and engaging. Do not alter or remove any fundamental aspect of the framework.
2. **Originality and algorithmic challenge**: Within the framework boundaries, design the problem so that it is non-trivial and requires meaningful algorithmic reasoning. The problem should encourage diverse solution strategies and subtle algorithmic thinking.
3. **No meta-comments or reasoning steps**: Exclude all internal deliberation, step-by-step reasoning, and self-referential phrases. Avoid using terms such as "wait" "let's" or any similar language that indicates thinking.
4. **Strict output format**: Produce only a fully completed Problem Template block. The model's entire output must be exactly the filled template, with no extra text or commentary outside the template.
5. **Well-specified constraints**: Ensure input/output and formal constraints are complete, and the intended algorithmic complexity is clear.

### Problem Framework
{framework}

### Problem Template
```
**Problem Title:**
[Your problem title]

**Tags:**
[Comma-separated topics/tags]

**Difficulty:**
[Easy / Medium / Hard / Very Hard]

**Problem Statement:**
[Description of the problem]

**Input:**
[Exact input format]

**Output:**
[Exact output format]

**Constraints:**
[Formal constraints with bounds]

**Example:**
[Examples without explanations]

**Notes:** (optional)
[A brief reference solution description]
```

Important: The output produced by you now must be ONLY the completed `Problem Template` block filled according to the above rules, without including any text unrelated to the algorithm problem itself.

### Your Designed Problem
\end{lstlisting}
\end{tcolorbox}

\captionof{figure}{Prompt Template for \textit{Step 3: Problem Synthesis} in ADR (algorithmic task).}
\label{fig:adr_prompt_algorithmic_step_3}

\begin{tcolorbox}[breakable, width=1\textwidth, fontupper=\small, colback=blue!2, boxrule=0.9pt] 
\begin{lstlisting}[breaklines=true, basicstyle=\ttfamily\small]
## Task
You are given an algorithm problem. Your task is to generate both the `solution code` and the `test case generator code` for that algorithm problem.

## Output format
<|Solution Begin|>
[Solution Code in Python]
<|Solution End|>
<|Test Case Generator Begin|>
[Test Case Generator in Python]
<|Test Case Generator End|>

## Example for `stdin_stdout` solution
<|Solution Begin|>
```python
import sys

def main():
    data = sys.stdin.read().strip().split()
    if len(data) < 2:
        return
    a, b = int(data[0]), int(data[1])
    print(a + b)

if __name__ == "__main__":
    main()
```
<|Solution End|>
<|Test Case Generator Begin|>
```python
import random

def generate_test_cases():
    random.seed(42)
    inputs = []
    outputs = []

    pairs = [
        (0, 0),
        (1, -1),
        (-1, -1),
        (10**6, 10**6),
        (-10**6, -10**6),
        (10**6, -10**6),
        (123456, 654321)
    ]

    NUM_RANDOM = 10
    MINV, MAXV = -10**6, 10**6
    for _ in range(NUM_RANDOM):
        a = random.randint(MINV, MAXV)
        b = random.randint(MINV, MAXV)
        pairs.append((a, b))

    for a, b in pairs:
        inputs.append(f"{{a}} {{b}}\n")
        outputs.append(f"{{a + b}}\n")

    return {{
        "input": inputs,
        "output": outputs,
        "fn_name": None,
        "type": "stdin_stdout"
    }}
```
<|Test Case Generator End|>

## Example for `function_call` solution
<|Solution Begin|>
```python
def add(a, b):
    return a + b
```
<|Solution End|>
<|Test Case Generator Begin|>
```python
import random

def generate_test_cases():
    random.seed(42)
    inputs = []
    outputs = []

    pairs = [
        (0, 0),
        (1, -1),
        (-1, -1),
        (10**6, 10**6),
        (-10**6, -10**6),
        (10**6, -10**6),
        (123456, 654321)
    ]

    NUM_RANDOM = 10
    MINV, MAXV = -10**6, 10**6
    for _ in range(NUM_RANDOM):
        a = random.randint(MINV, MAXV)
        b = random.randint(MINV, MAXV)
        pairs.append((a, b))

    for a, b in pairs:
        inputs.append([a, b])
        outputs.append([a + b])

    return {{
        "input": inputs,
        "output": outputs,
        "fn_name": "add",
        "type": "function_call"
    }}
```
<|Test Case Generator End|>

## Algorithm problem
{problem}
\end{lstlisting}
\end{tcolorbox}

\captionof{figure}{Prompt Template for \textit{Step 4: Execution-grounded Validation} in ADR (algorithmic task).}
\label{fig:adr_prompt_algorithmic_step_4}

\begin{tcolorbox}[breakable, width=1\textwidth, fontupper=\small, colback=blue!2, boxrule=0.9pt] 
\begin{lstlisting}[breaklines=true, basicstyle=\ttfamily\small]
You are given a programming problem and a reference correct solution.

Your task is to generate exactly 5 distinct **near-miss solutions**.

A near-miss solution is:
- Logically plausible and well-structured
- Likely to pass many simple or random test cases
- Incorrect due to subtle flaws (e.g., edge cases, off-by-one errors, incorrect assumptions, partial logic, numerical instability, complexity limits)

### Instructions:
1. Each near-miss solution should be written as full executable code.
2. Each solution must fail for at least one non-trivial or adversarial input.
3. The mistakes should be diverse. Avoid repeating the same error pattern.
4. Do NOT explicitly state what the bug is in the code.
5. Do NOT include explanations inside the code.
6. Output the solutions as a numbered list from 1 to 5.

### Problem description:
{problem}

### Correct reference solution:
```python
{reference_solution}
```
\end{lstlisting}
\end{tcolorbox}

\captionof{figure}{Prompt Template for \textit{Step 5: Adversarial Solution Space Refinement} in ADR.}
\label{fig:adr_adversarial_step_1}

\begin{tcolorbox}[breakable, width=1\textwidth, fontupper=\small, colback=blue!2, boxrule=0.9pt] 
\begin{lstlisting}[breaklines=true, basicstyle=\ttfamily\small]
You are given:
- A programming problem
- A correct reference solution
- A set of near-miss solutions that are incorrect but pass many naive tests
- The current version of a test case generator

Your task is to improve the test case generator using **adversarial reasoning**.

### Objective:
- Modify or redesign the test case generator to **maximize the failure rate of the near-miss solutions**
- While ensuring that the correct reference solution still passes all generated test cases

### Instructions:
1. Analyze the common and uncommon weaknesses likely present in the near-miss solutions.
2. Design test cases that specifically target:
   - Edge cases
   - Boundary conditions
   - Rare corner scenarios
   - Stress limits (size, value ranges, ordering, structure)
   - Implicit assumptions likely made by incorrect solutions
3. The generator should be general and reusable, not hardcoded for a single bug.
4. Do NOT explicitly reference individual near-miss solutions in the generator logic.
5. Output the improved test case generator as executable code or clear pseudocode.

### Problem description:
{problem}

### Correct reference solution:
```python
{reference_solution}
```

### Near-miss solutions:
{near_miss_solutions}

### Current test case generator:
```python
{test_case_generator}
```
\end{lstlisting}
\end{tcolorbox}

\captionof{figure}{Prompt Template for \textit{Step 5: Adversarial Solution Space Refinement} in ADR.}
\label{fig:adr_adversarial_step_2}

\begin{tcolorbox}[breakable, width=1\textwidth, fontupper=\small, colback=blue!2, boxrule=0.9pt] 
\begin{lstlisting}[breaklines=true, basicstyle=\ttfamily\small]
Please analyze the following tool-calling code problem according to the guidelines below. The output should follow the format below (do not add any other notes or explanations):
<answer>
Computational Objective:
Tool Dependency Set:
Processing Logic Constraints:
Input Interface:
Output Specification:
</answer>

### Guidelines
1. Computational Objective
    * **Definition:** An abstract specification of the primary computation or transformation to be performed on data, independent of implementation details.
    * **Role:** Defines *what* the task fundamentally does; without it, the task has no semantic goal.
    * **Variation Axes:**
        * Type of computation (aggregation, generation, transformation, analysis)
        * Deterministic vs stochastic behavior
        * Single-stage vs multi-stage computation

2. Tool Dependency Set
    * **Definition:** The set of external libraries, modules, or tools that must be imported and used to accomplish the task.
    * **Role:** Distinguishes this task type as a *tool-calling* problem; ensures the task demonstrates use of specific utilities beyond core language constructs.
    * **Variation Axes:**
        * Standard library vs third-party tools
        * Number of tools
        * Functional role of tools (randomness, statistics, iteration, collections, etc.)

3. Processing Logic Constraints
    * **Definition:** High-level rules or required operations that constrain how the computation must be carried out, without prescribing exact code.
    * **Role:** Ensures the task exercises particular patterns (e.g., shuffling before computing, sorting by a derived metric).
    * **Variation Axes:**
        * Ordering of operations
        * Required intermediate transformations
        * Use of specific functions or methods from the tools

4. Input Interface
    * **Definition:** A formal description of the function inputs, including parameter names, types, defaults, and constraints.
    * **Role:** Establishes how external data enters the task; necessary for invoking the computation.
    * **Variation Axes:**
        * Number of parameters
        * Data types (scalars, lists, dicts, strings, etc.)
        * Default values
        * Validity constraints (ranges, non-negativity, non-emptiness)

5. Output Specification
    * **Definition:** A precise description of the expected output type, structure, and semantic meaning.
    * **Role:** Defines task completion criteria; without it, correctness cannot be evaluated.
    * **Variation Axes:**
        * Output data type (float, dict, list, etc.)
        * Structural properties (sorted, aggregated, keyed by...)
        * Deterministic vs stochastic output interpretation

### Tool-Calling Code Problem
{problem}

### Solution of the Tool-Calling Code Problem
```python
{solution}
```

### Analysis
\end{lstlisting}
\end{tcolorbox}

\captionof{figure}{Prompt Template for \textit{Step 1: Element Extraction} in ADR (tool usage task).}
\label{fig:adr_prompt_tool_step_1}

\begin{tcolorbox}[breakable, width=1\textwidth, fontupper=\small, colback=blue!2, boxrule=0.9pt] 
\begin{lstlisting}[breaklines=true, basicstyle=\ttfamily\small]
You are an expert in tool-calling code task abstraction and schema-level task design.
You are given:
- One randomly sampled Computational Objective
- One randomly sampled Tool Dependency Set
- Three reference sets, each consisting of:
    - Processing Logic Constraints
    - Input Interface
    - Output Specification
Your task is to design a completely new tool-calling code task at the element level, by inferring a novel and coherent combination of task elements, not by copying or minimally editing the references.
The output should follow the format below (do not add any other notes or explanations):
<answer>
Computational Objective:
Tool Dependency Set:
Processing Logic Constraints:
Input Interface:
Output Specification:
</answer>

### Core Requirements

1. **Element-level generation only**
    * Do NOT write a concrete problem statement or code.
    * Do NOT reuse wording, structure, or semantics from any single reference set.
    * Operate strictly at the level of abstract task elements.

2. **Five-element completeness**
    You must generate **exactly five elements**, one for each of the following:
    * Computational Objective
    * Tool Dependency Set
    * Processing Logic Constraints
    * Input Interface
    * Output Specification

3. **Consistency constraints**
    * The Computational Objective must be **achievable** using the Tool Dependency Set.
    * Processing Logic Constraints must **meaningfully constrain** how the tools are used.
    * Input Interface must provide **sufficient information** to execute the objective.
    * Output Specification must be a **direct consequence** of the objective and logic.

4. **Novel recombination**
    * Treat the three reference sets as **design signals**, not templates.
    * The resulting element set should be plausibly generatable by recombining ideas,
      but **must not align exactly with any reference along more than one element**.

5. **Tool-calling emphasis**
    * The Tool Dependency Set must play a **non-trivial role** in enabling or shaping the task.
    * If tools were removed, the task should lose its defining character.

### Given Computational Objective
{computational_objective}

### Given Tool Dependency Set
{tool_dependency_set}

### Reference Sets
{combinations_1}

{combinations_2}

{combinations_3}

### New Elements
\end{lstlisting}
\end{tcolorbox}

\captionof{figure}{Prompt Template for \textit{Step 2: Controlled Recombination} in ADR (tool usage task).}
\label{fig:adr_prompt_tool_step_2}

\begin{tcolorbox}[breakable, width=1\textwidth, fontupper=\small, colback=blue!2, boxrule=0.9pt] 
\begin{lstlisting}[breaklines=true, basicstyle=\ttfamily\small]
You are an expert in **tool-calling code task synthesis**.

You are given a complete task specification expressed as **five abstract task elements**:

* Computational Objective
* Tool Dependency Set
* Processing Logic Constraints
* Input Interface
* Output Specification

Your task is to **synthesize a single, complete tool-calling code problem** that strictly follows the required template and **faithfully instantiates all five elements**.

### Core Rules
1. **Element Fidelity**
    * All five elements must be fully and explicitly reflected.
    * Do not add, remove, or reinterpret any requirement.

2. **No Meta or Reasoning Text**
    * Output only a finished problem statement.
    * No explanations, hints, or commentary.

3. **Mandatory Tool Usage**
    * The task must inherently require the specified tools.
    * Removing the tools should break the task.

4. **Template Exactness**
    * Follow the template structure and section order exactly.
    * Do not rename, reorder, or omit sections.

5. **Header Code Authority**
    * Use the header code verbatim.
    * All required imports and the function signature must appear there, and only there.

### Output Template
**Problem Description:**
<Concise natural-language description instantiating the Computational Objective and Processing Logic Constraints>

**Input:**
<Formal description derived from the Input Interface>

**Output:**
<Formal description derived from the Output Specification>

**Constraints & Requirements:**
* <Each Processing Logic Constraint as a concrete requirement>
* <Any implicit constraints required by the Tool Dependency Set>

**Header Code:**
```python
<header_code>
```

### Given Elements
{elements}

### Your Designed Problem
\end{lstlisting}
\end{tcolorbox}

\captionof{figure}{Prompt Template for \textit{Step 3: Problem Synthesis} in ADR (tool usage task).}
\label{fig:adr_prompt_tool_step_3}

\begin{tcolorbox}[breakable, width=1\textwidth, fontupper=\small, colback=blue!2, boxrule=0.9pt] 
\begin{lstlisting}[breaklines=true, basicstyle=\ttfamily\small]
## Task
You are given a tool-calling code problem. Your task is to generate both the `solution code` and the `test code` in pytest for that problem.

## Output format
<|Solution Begin|>
[Solution Code in Python]
<|Solution End|>
<|Test Code Begin|>
[Test Code in Pytest]
<|Test Code End|>

## Example
<|Solution Begin|>
```python
import math

def add(a, b):
    \"\"\"
    Return the sum of a and b.
    \"\"\"
    return math.fsum([a, b])
```
<|Solution End|>
<|Test Code Begin|>
```python
# Sanity Check
def test_add_basic():
    assert add(1, 2) == 3.0

# Edge Cases
def test_add_with_zero():
    assert add(0, 5) == 5.0
    assert add(0, 0) == 0.0

def test_add_negative_numbers():
    assert add(-1, -2) == -3.0
    assert add(-1, 1) == 0.0

# Extreme Cases
def test_add_large_numbers():
    large = 10**18
    assert add(large, large) == float(2 * large)

def test_add_float_precision():
    # math.fsum should handle precision better than naive addition
    result = add(1e16, 1.0)
    assert result == math.fsum([1e16, 1.0])

# Boundary of Input Domain
def test_add_mixed_int_float():
    assert add(1, 2.5) == 3.5
```
<|Test Code End|>

## Tool-Calling Code Problem
{problem}
\end{lstlisting}
\end{tcolorbox}

\captionof{figure}{Prompt Template for \textit{Step 4: Execution-grounded Validation} in ADR (tool usage task).}
\label{fig:adr_prompt_tool_step_4}

\begin{tcolorbox}[breakable, width=1\textwidth, fontupper=\small, colback=blue!2, boxrule=0.9pt] 
\begin{lstlisting}[breaklines=true, basicstyle=\ttfamily\small]
Please analyze the following data science problem according to the guidelines below. The output should follow the format below (do not add any other notes or explanations):
<answer>
Data Schema:
Task Goal:
Output Contract:
Implementation Environment:
Behavioral Constraints:
</answer>

### Guidelines
1. Data Schema
    * **Definition:** A precise, abstract description of the input data the code will consume: its origin (file, DB, in-memory), high-level structure (table/array/tensor/graph), dimensions or shape signatures, named fields or column types, and any distributional or ordering properties that affect processing.
    * **Role:** States what the program receives so implementers know how to parse, index, and validate inputs; without it the operation cannot be applied correctly.
    * **Variation Axes:** source (CSV/JSON/SQL/in-memory), container type (DataFrame / ndarray / sparse matrix / list / dict), schema detail level (full typed schema vs. loose shape), size scale (small/medium/large/streaming), ordering guarantees (sorted/partitioned/grouped/random).

2. Task Goal
    * **Definition:** A concise, testable description of the transformation, computation, or analysis to perform on the input data. It describes the expected semantic outcome (e.g., "reorder rows by index list", "normalize each column in-place", "reshape a 1-D sequence into a 2-D matrix").
    * **Role:** Provides the objective that determines what code must do; without it there is no target behavior to implement.
    * **Variation Axes:** transformation type (filter/aggregate/transform/reorder/reshape/compute stat), granularity (column-wise/row-wise/matrix-level/element-wise), in-place vs. functional (mutating or returning new object), deterministic vs. randomized, single-step vs. pipeline of steps.

3. Output Contract
    * **Definition:** An explicit specification of the result: types, shapes, naming/variable expectations, ordering of outputs, side-effects (files written, in-place mutation), and the acceptance criteria (what constitutes a correct result).
    * **Role:** Defines how the success of the Task Goal is observed and integrated downstream; it removes ambiguity about return types and side-effects.
    * **Variation Axes:** return style (value/tuple/None + side-effects), naming (specific variable name required vs. any), strictness of shape/type (exact shape vs. compatible), required persistence (save to file/DB vs. ephemeral), error signaling mode (exceptions/error codes).

4. Implementation Environment
    * **Definition:** The technical context in which the solution must run: permitted libraries and APIs.
    * **Role:** Constrains feasible solutions and guides use of tool-specific APIs; without it solutions may use unsupported constructs or libraries.
    * **Variation Axes:** allowed libraries and formats.

5. Behavioral Constraints
    * **Definition:** Non-functional and precondition requirements that affect algorithm choice: mutability requirements, numerical stability, memory/time complexity targets, guaranteed invariants in inputs (e.g., "columns contain non-negative numbers"), and failure modes to avoid.
    * **Role:** Ensures the implementation respects important operational and correctness constraints that the Task Goal alone does not cover.
    * **Variation Axes:** mutability (must/should not mutate input), complexity bounds (O(n)/O(n log n)/approximate), numeric precision needs (float32/64, tolerance), concurrency safety, expected input cleanliness (may contain NaNs/missing/duplicates).

### Data Science Code Problem
{problem}

### Solution of the Data Science Code Problem
```python
{solution}
```

### Analysis
\end{lstlisting}
\end{tcolorbox}

\captionof{figure}{Prompt Template for \textit{Step 1: Element Extraction} in ADR (data science task).}
\label{fig:adr_prompt_data_step_1}

\begin{tcolorbox}[breakable, width=1\textwidth, fontupper=\small, colback=blue!2, boxrule=0.9pt] 
\begin{lstlisting}[breaklines=true, basicstyle=\ttfamily\small]
You are an expert in data science code task abstraction and schema-level task design.
You are given:
- One randomly sampled Task Goal
- One randomly sampled Data Schema
- Three reference sets, each consisting of:
    - Output Contract
    - Implementation Environment
    - Behavioral Constraints
Your task is to design a completely new data science code task at the element level, by inferring a novel and coherent combination of task elements, not by copying or minimally editing the references.
The output should follow the format below (do not add any other notes or explanations):
<answer>
Data Schema:
Task Goal:
Output Contract:
Implementation Environment:
Behavioral Constraints:
</answer>

### Core Requirements

1. **Element-level generation only**
    * Do NOT write a concrete problem statement or code.
    * Do NOT reuse wording, structure, or semantics from any single reference set.
    * Operate strictly at the level of abstract task elements.

2. **Five-element completeness**
    You must generate **exactly five elements**, one for each of the following:
    * Data Schema
    * Task Goal
    * Output Contract
    * Implementation Environment
    * Behavioral Constraints

3. **Consistency constraints**
    * Any Task Goal is valid with any Data Schema provided the Output Contract and Implementation Environment are consistent to ensure API compatibility (e.g., a matrix normalization goal must pair with a container type that supports columns).
    * Behavioral Constraints must be compatible with Implementation Environment (e.g., a constraint requiring in-place mutation is invalid if environment disallows mutating APIs).

4. **Novel recombination**
    * Treat the three reference sets as **design signals**, not templates.
    * The resulting element set should be plausibly generatable by recombining ideas,
      but **must not align exactly with any reference along more than one element**.
    * Different Task Goals can be legally combined with different Data Schemas (e.g., change from pandas DataFrame to Spark DataFrame -> update environment and acceptable complexity) (e.g., from "reorder rows" -> "group and aggregate by column").
    * Produce variants by changing Variation Axes independently: change allowed libraries in Implementation Environment while keeping Data Schema and Task Goal identical; the Output Contract and Examples must be updated only if necessary.

5. **Data science emphasis**
    * The task must be inherently data-centric (e.g., transformation, aggregation, statistical computation, reshaping, validation).
    * If the Data Schema were removed or altered, the task should lose its defining meaning.

### Given Data Schema
{data_schema}

### Given Task Goal
{task_goal}

### Reference Sets
{combinations_1}

{combinations_2}

{combinations_3}

### New Elements
\end{lstlisting}
\end{tcolorbox}

\captionof{figure}{Prompt Template for \textit{Step 2: Controlled Recombination} in ADR (data science task).}
\label{fig:adr_prompt_data_step_2}

\begin{tcolorbox}[breakable, width=1\textwidth, fontupper=\small, colback=blue!2, boxrule=0.9pt] 
\begin{lstlisting}[breaklines=true, basicstyle=\ttfamily\small]
You are an expert in **data science code task synthesis**.

You are given a complete task specification expressed as **five abstract task elements**:

* Data Schema
* Task Goal
* Output Contract
* Implementation Environment
* Behavioral Constraints

Your task is to **synthesize a single, complete data science code problem** that strictly follows the required template and **faithfully instantiates all five elements**.

### Core Rules
1. **Element Fidelity**
    * All five elements must be fully and explicitly reflected.
    * Do not add, remove, or reinterpret any requirement.

2. **No Meta or Reasoning Text**
    * Output only a finished problem statement.
    * No explanations, hints, or commentary.

3. **Mandatory Implementation Environment Usage**
    * The task must inherently require the specified implementation environment.
    * Removing the tools should break the task.

4. **Template Exactness**
    * Follow the template structure and section order exactly.
    * Do not rename, reorder, or omit sections.

5. **Header Code Authority**
    * Use the header code verbatim.
    * All required imports and the function signature must appear there, and only there.

### Output Template
**Problem Description:**
<Concise natural-language description instantiating the Computational Objective and Processing Logic Constraints>

**Input:**
<Formal description derived from the Input Interface>

**Output:**
<Formal description derived from the Output Specification>

**Constraints & Requirements:**
* <Each Processing Logic Constraint as a concrete requirement>
* <Any implicit constraints required by the Tool Dependency Set>

**Header Code:**
```python
<header_code>
```

### Given Elements
{elements}

### Your Designed Problem
\end{lstlisting}
\end{tcolorbox}

\captionof{figure}{Prompt Template for \textit{Step 3: Problem Synthesis} in ADR (data science task).}
\label{fig:adr_prompt_data_step_3}

\begin{tcolorbox}[breakable, width=1\textwidth, fontupper=\small, colback=blue!2, boxrule=0.9pt] 
\begin{lstlisting}[breaklines=true, basicstyle=\ttfamily\small]
You are an expert in **data science code task synthesis**.

You are given a complete task specification expressed as **five abstract task elements**:

* Data Schema
* Task Goal
* Output Contract
* Implementation Environment
* Behavioral Constraints

Your task is to **synthesize a single, complete data science code problem** that strictly follows the required template and **faithfully instantiates all five elements**.

### Core Rules
1. **Element Fidelity**
    * All five elements must be fully and explicitly reflected.
    * Do not add, remove, or reinterpret any requirement.

2. **No Meta or Reasoning Text**
    * Output only a finished problem statement.
    * No explanations, hints, or commentary.

3. **Mandatory Implementation Environment Usage**
    * The task must inherently require the specified implementation environment.
    * Removing the tools should break the task.

4. **Template Exactness**
    * Follow the template structure and section order exactly.
    * Do not rename, reorder, or omit sections.

5. **Header Code Authority**
    * Use the header code verbatim.
    * All required imports and the function signature must appear there, and only there.

### Output Template
**Problem Description:**
<Concise natural-language description instantiating the Computational Objective and Processing Logic Constraints>

**Input:**
<Formal description derived from the Input Interface>

**Output:**
<Formal description derived from the Output Specification>

**Constraints & Requirements:**
* <Each Processing Logic Constraint as a concrete requirement>
* <Any implicit constraints required by the Tool Dependency Set>

**Header Code:**
```python
<header_code>
```

### Given Elements
{elements}

### Your Designed Problem
\end{lstlisting}
\end{tcolorbox}

\captionof{figure}{Prompt Template for \textit{Step 4: Execution-grounded Validation} in ADR (data science task).}
\label{fig:adr_prompt_data_step_4}

\begin{tcolorbox}[breakable, width=1\textwidth, fontupper=\small, colback=blue!2, boxrule=0.9pt] 
\begin{lstlisting}[breaklines=true, basicstyle=\ttfamily\small]
You are an expert in code task abstraction and problem schema design.

You are given:
1. A specified code task type
2. Three concrete example tasks of this type

Your task is to infer a **minimal and sufficient set of composable task elements** such that:
- Using only these elements, one can construct a **complete and valid code task**
- The elements are **orthogonal** (each captures a distinct aspect of the task)
- The elements support **recombination and variation**, enabling the generation of new tasks of the same type

Follow these strict requirements:

### Step 1: Element Discovery
Identify the **smallest possible set of task elements** that are:
- Necessary (removing any one would make the task incomplete)
- General (not tied to any specific example)
- Reusable across different tasks of the same type

### Step 2: Element Definition
For each element, provide:
- **Element Name** (concise, canonical)
- **Element Definition**: a precise, abstract description of what this element represents
- **Role in Task Construction**: why this element is required to form a complete task
- **Variation Axes**: what aspects of this element can vary to generate different tasks

Definitions must be **task-agnostic templates**, not filled instances.

### Step 3: Element Interaction Rules
Describe:
- Which elements are mandatory vs optional
- Valid combinations and ordering constraints (if any)
- How different elements can be recombined without breaking task validity

### Output Format (Strict)
Return the result in the following structure:

```
Task Type: ...

Minimal Task Element Set:
1. Element Name
   - Definition:
   - Role:
   - Variation Axes:
2. ...

Element Interaction & Composition Rules:
- ...
```

### Given Code Task Type
{task_type}

### Given Code Task Demonstrations
{example_1}

{example_2}

{example_3}
\end{lstlisting}
\end{tcolorbox}

\captionof{figure}{Prompt Template for \textit{Info-Guided Element Schema Optimization (initialize element schema)} in ADR.}
\label{fig:adr_optimization_1}

\begin{tcolorbox}[breakable, width=1\textwidth, fontupper=\small, colback=blue!2, boxrule=0.9pt] 
\begin{lstlisting}[breaklines=true, basicstyle=\ttfamily\small]
You are an expert in {task_type} design and representation learning.

Your task is to improve the ELEMENT SCHEMA used to synthesize {task_type} problems.

### Current element schema
{old_element_schema}

### Your goal
The Schema has been decomposed into a set of elements, and for each element you are provided with:
- The entropy of the element (measuring its information content and uncertainty)
- The conditional mutual information (CMI) between pairs of elements with respect to the programming problem (measuring redundancy, dependency, or complementary information)

Your task is to analyze and optimize the initial element Schema based on information-theoretic principles.

You may perform the following actions:
1. ADD a new element
2. REMOVE an existing element
3. MERGE two or more elements
4. SPLIT one element into multiple elements
5. REFINE the definition of an element

### The entropy of the element
{element_schema_entropy}

### The CMI between pairs of elements
{element_schema_cmi}

### Important constraints
- The given schema may contain several elements, but make sure the final schema should NOT exceed **5** elements.
- Each element must be:
  - Clearly defined
  - Usable in a problem generation prompt
  - Non-redundant
- Prefer elements that:
  - Reduce ambiguity of the optimal solution
  - Improve difficulty controllability
  - Increase diversity of generated problems

### Output format (strict JSON)
{{
  "change_proposals": [
    {{
      "action": "add | remove | merge | split | refine",
      "target_elements": ["ElementA", "ElementB"],
      "new_elements": [
        {{
          "name": "...",
          "definition": "..."
        }}
      ],
      "rationale": "Why this change improves problem quality"
    }}
  ]
}}
\end{lstlisting}
\end{tcolorbox}

\captionof{figure}{Prompt Template for \textit{Info-Guided Element Schema Optimization (optimize schema based on the information theory metrics)} in ADR.}
\label{fig:adr_optimization_2}


\end{document}